\pdfoutput=1
\documentclass[10pt]{article}

\usepackage[preprint]{tmlr}
\newif\ifpreprint
\preprinttrue

\usepackage{amsmath,amssymb}
\usepackage{booktabs}
\usepackage{graphicx}
\usepackage{microtype}
\usepackage{multirow}
\usepackage{array}
\usepackage{xcolor}
\usepackage{xurl}
\usepackage{hyperref}

\title{Calibration Language Shapes Block-Influence Pruning in Multilingual Language Models}

\author{\name Vedant Pandhare \email vedant11734@gmail.com \\
      \addr Independent Researcher
      \AND
      \name Piyush Dhoka \email piyushdhoka007@gmail.com \\
      \addr Independent Researcher}

\newcommand{\coderepo}{\url{https://github.com/VedantPandhare/Language-Balanced-Depth-Pruning-of-Small-Multilingual-LLMs}}

\newcommand{\BI}{\mathrm{BI}}
\newcommand{\EN}{\textsc{en}}
\newcommand{\HI}{\textsc{hi}}
\newcommand{\MR}{\textsc{mr}}
\newcommand{\TA}{\textsc{ta}}

\newcommand{\PPL}{\mathrm{PPL}}
\newcommand{\Rtwo}{\mathrm{RL2}}

\newcommand{\code}[1]{\texttt{#1}}

\begin{document}

\maketitle

\begin{abstract}
Block Influence (BI) ranks Transformer blocks for depth pruning from their input--output cosine change, but the ranking depends on the calibration distribution. We study whether English calibration transfers to Indian languages in Qwen2.5-1.5B-Instruct, Llama-3.2-1B, and Gemma-2-2B. English, Hindi, Marathi, and Tamil use disjoint Wikipedia articles for calibration and evaluation; every selector receives the same 150-chunk budget, and Tamil is excluded from multilingual aggregation to test cross-script transfer. Across all three models, late-block BI is 20.7--55.5\% higher for the three Indian languages than for English, and every predeclared late-minus-middle contrast has a 95\% bootstrap interval above zero. This mismatch changes the low-BI blocks selected for removal despite high global rank correlations. Averaging BI over equal English, Hindi, and Marathi budgets mitigates the resulting perplexity damage for Qwen at six removed blocks and for Gemma at all tested depths, including held-out Tamil. The mitigation is not universal: on Llama, balancing improves Hindi and Marathi at two removed blocks but severely harms Tamil, while larger pruning budgets select the same blocks and collapse perplexity regardless of calibration. Rank-mean and minimax-rank aggregators do not dominate the raw mean. These results support a diagnostic claim rather than a universal pruning rule: English-only BI systematically misses late-layer activity for Indian-language text, but multilingual calibration must still be validated per architecture and pruning budget.
\end{abstract}

\section{Introduction}
\label{sec:introduction}

Removing complete Transformer blocks is an attractive compression operation: it shortens the sequential computation graph, reduces parameters, and does not require sparse kernels. Block-pruning methods therefore need to answer a consequential discrete question: which blocks can be removed? Block Influence (BI), introduced by ShortGPT, answers it by measuring how little each block changes its hidden states on a small calibration corpus and pruning the blocks with the lowest scores \citep{men2024shortgpt}. The score is inexpensive and training-free, but it is conditional on the corpus used to estimate it.

That condition matters for multilingual deployment. Calibration data in language-model compression is commonly English even when the model will serve other languages \citep{kurz2026limitations,zeng2024mbs}. Prior work shows calibration-language effects for unstructured weight pruning and quantization \citep{kurz2026limitations,choenni2025mwanda}, but those results do not establish whether the \emph{ordering of whole Transformer blocks} transfers across languages. A small score perturbation is sufficient to change a depth-pruning decision, and removing one complete block has a much larger structural effect than changing an equal number of isolated weights.

We examine this problem for English (\EN), Hindi (\HI), Marathi (\MR), and Tamil (\TA). Hindi and Marathi provide related Devanagari-script languages, while Tamil is from a different language family and uses a different script. We include Tamil in analysis and evaluation but exclude it from language-balanced selector construction. This design asks whether a selector balanced over \EN/\HI/\MR{} transfers beyond the scripts and languages it observes, rather than merely fitting the calibration mixture.

Our central claim is deliberately limited:
\begin{quote}
\emph{English-only calibration systematically underestimates late-block influence for the Indian-language text studied here. Language-balanced calibration can avoid harmful English-selected blocks, but the downstream benefit is architecture- and budget-dependent.}
\end{quote}
The first sentence is a statement about the behavior of BI. The second is an empirical qualification, not a claim that balancing guarantees safe pruning.

We test this claim on three architecture families: Qwen2.5-1.5B-Instruct, Llama-3.2-1B, and Gemma-2-2B. The design uses article-disjoint calibration and evaluation data, equal token budgets, five independently sampled calibration subsets, three aggregation rules, ten random-pruning controls, and uncertainty over both calibration subsets and held-out evaluation articles. No recovery training is used; this isolates block selection from post-pruning adaptation.

The study makes four contributions:
\begin{enumerate}
    \item We provide a controlled analysis of calibration-language transfer for BI-based \emph{whole-block} pruning, complementing prior multilingual studies of unstructured pruning.
    \item We localize the cross-language discrepancy: across all three models and all three Indian languages, the target-to-English BI ratio is substantially larger in a predeclared late-depth region than in the middle region.
    \item We compare equal-budget raw-mean, rank-mean, and minimax-rank multilingual selectors, including transfer to held-out Tamil. Raw-mean balancing prevents damaging late-block choices in Qwen and Gemma, but no aggregator dominates on every model.
    \item We report a negative boundary case. BI pruning is brittle on the 16-block Llama model, and the balanced selector's Hindi/Marathi gains at one budget do not transfer to Tamil. This separates detecting calibration mismatch from correcting it.
\end{enumerate}

The results suggest a practical rule of evidence: multilingual depth pruning should report language-conditioned block scores and held-out-language consequences, not infer robustness from an English calibration set or from a high global rank correlation alone.

\section{Related Work}
\label{sec:related}

\paragraph{Whole-block pruning.}
ShortGPT defines BI from block input--output cosine similarity and removes low-influence Transformer blocks without training \citep{men2024shortgpt}. Other depth-reduction methods use output reconstruction, redundancy verification, block merging, or a short recovery stage \citep{song2024sleb,yang2024laco,gromov2025deep}. These methods show that complete-block removal can reduce sequential computation, but evaluations and calibration corpora are predominantly English. Our work does not introduce another recovery procedure. It examines a prior question: whether BI identifies the same low-influence blocks when the calibration language changes.

\paragraph{Calibration data for compression.}
SparseGPT and Wanda made small calibration sets standard in post-training weight pruning \citep{frantar2023sparsegpt,sun2024wanda}. Calibration content can materially alter pruning outcomes even within English \citep{bandari2024c4}. Multilingual Brain Surgeon samples several languages to mitigate English-centric compression \citep{zeng2024mbs}; bilingual pruning \citep{kim2024multilingual} and M-Wanda \citep{choenni2025mwanda} likewise incorporate cross-lingual information. Most directly, \citet{kurz2026limitations} compare target-language calibration for Wanda and SparseGPT over seven languages. They find consistent perplexity benefits but weaker downstream-task transfer. Those studies concern masks over individual weights or neurons. We instead study the discrete ranking and removal of whole decoder blocks under BI.

\paragraph{Multilingual structured pruning.}
\citet{wibowo2025iterative} study structured layer pruning together with iterative recovery and distillation, showing that the recovery language has nontrivial cross-lingual effects. Our setting is complementary: one-shot block removal with fixed weights allows the language dependence of \emph{selection} to be observed without adaptation. Work on multilingual model compression more broadly reports disproportionate damage to some languages and scripts \citep{ogueji2022compression,petrov2023tokenizers}.

\paragraph{Depth and multilingual representation.}
Analyses of multilingual Transformers find both shared representations and language-specific components, with language identity and surface realization often becoming prominent toward the boundaries of the network \citep{tang2024neurons,wendler2024llamas,zhao2024multilingualism}. This motivates, but does not imply, our late-layer hypothesis. We test it directly using a predeclared depth partition and independent calibration articles.

\paragraph{Positioning.}
We therefore do not claim that calibration-language sensitivity is new in all forms of pruning. The novelty is the combination of (i) BI whole-block ranking, (ii) localization across normalized depth, (iii) equal-budget multilingual aggregation ablations, and (iv) an Indian-language evaluation where Tamil is held out from aggregation. The Llama counterexample also provides evidence about when a calibration-aware selector is insufficient.

\section{Method}
\label{sec:method}

\subsection{Language-conditioned Block Influence}

Consider a decoder with blocks $l\in\{0,\ldots,L-1\}$. Let $\mathbf{h}_{l,t}^{\mathrm{in}}$ and $\mathbf{h}_{l,t}^{\mathrm{out}}$ be the input and output hidden states of block $l$ at token position $t$. For a calibration collection $D_\lambda$ in language $\lambda$, we estimate
\begin{equation}
\BI_l^{(\lambda)}
=1-\frac{1}{|D_\lambda|}
\sum_{x\in D_\lambda}\frac{1}{|\mathcal{T}_x|}
\sum_{t\in\mathcal{T}_x}
\frac{
(\mathbf{h}_{l,t}^{\mathrm{in}})^\top\mathbf{h}_{l,t}^{\mathrm{out}}
}{
\|\mathbf{h}_{l,t}^{\mathrm{in}}\|_2
\|\mathbf{h}_{l,t}^{\mathrm{out}}\|_2
}.
\label{eq:bi}
\end{equation}
The beginning-of-sequence position, when present, is excluded from $\mathcal{T}_x$ because it is an activation outlier and attention sink \citep{xiao2024streaming}. We capture both tensors with hooks on the block itself, before any final model normalization. Low BI means that a block changes the direction of its input representation little on that calibration distribution; it is a pruning candidate, not proof of mathematical redundancy.

For a score vector $\mathbf{s}$, $S_k(\mathbf{s})$ is the set of the $k$ lowest-scoring blocks, with block index as a deterministic final tie-breaker. We evaluate monolingual score vectors for \EN, \HI, \MR, and \TA. The first three are also used to construct language-balanced selectors; Tamil is never used in those aggregations.

\subsection{Equal-budget language-balanced selectors}

Let $\Lambda_{\mathrm{agg}}=\{\EN,\HI,\MR\}$. We compare three aggregators:
\begin{align}
s_l^{\mathrm{mean}}
  &= \frac{1}{|\Lambda_{\mathrm{agg}}|}
     \sum_{\lambda\in\Lambda_{\mathrm{agg}}}\BI_l^{(\lambda)}, \label{eq:rawmean}\\
s_l^{\mathrm{rank}}
  &= \frac{1}{|\Lambda_{\mathrm{agg}}|}
     \sum_{\lambda\in\Lambda_{\mathrm{agg}}}r_l^{(\lambda)}, \label{eq:rankmean}\\
s_l^{\mathrm{minimax}}
  &= \max_{\lambda\in\Lambda_{\mathrm{agg}}}r_l^{(\lambda)}, \label{eq:minimax}
\end{align}
where $r_l^{(\lambda)}\in[0,1]$ is the ascending percentile rank of block $l$ under language $\lambda$. Minimizing Eq.~\eqref{eq:minimax} prefers a block only when its rank is low for every aggregation language. Ties are broken by mean rank and then block index. Raw averaging retains score magnitudes; the rank rules remove cross-language scale differences.

Every primary selector sees 150 chunks. A monolingual selector receives 150 chunks from one language, while each balanced selector receives 50 chunks from each of \EN, \HI, and \MR. Thus any advantage cannot be attributed to giving balanced calibration three times as many tokens.

\subsection{Late-layer localization}
\label{sec:late-method}

We predeclare a depth partition before comparing outcomes. The final output-adjacent block is excluded because its role and BI scale are atypical. The remaining indices $0,\ldots,L-2$ are split as evenly as possible into early, middle, and late thirds. For target language $\lambda\in\{\HI,\MR,\TA\}$, define the layer-wise log ratio
\begin{equation}
q_l^{(\lambda)}
=\log\BI_l^{(\lambda)}-\log\BI_l^{(\EN)}.
\label{eq:logratio}
\end{equation}
Our localization statistic is
\begin{equation}
C^{(\lambda)}
=\frac{1}{|\mathcal{L}_{\mathrm{late}}|}
\sum_{l\in\mathcal{L}_{\mathrm{late}}}q_l^{(\lambda)}
-\frac{1}{|\mathcal{L}_{\mathrm{mid}}|}
\sum_{l\in\mathcal{L}_{\mathrm{mid}}}q_l^{(\lambda)}.
\label{eq:latecontrast}
\end{equation}
Positive $C^{(\lambda)}$ means that target-to-English influence is more elevated late in the network than in its middle. We also report the geometric percentage shift $100[\exp(\overline{q}_{\mathrm{late}})-1]$.

\subsection{Pruning outcome}

For each selected set $S$, we physically bypass those blocks without changing any remaining weight. On evaluation language $\ell$, mean token negative log-likelihood (NLL) gives
\begin{equation}
\PPL_\ell(S)=\exp\!\left(\frac{1}{N_\ell}
\sum_{i=1}^{N_\ell}-\log p_{\theta\setminus S}(x_i\mid x_{<i})\right).
\end{equation}
We report relative degradation against the intact model and compare English-only with balanced selection using
\begin{equation}
R_{\ell,k}
=\frac{\PPL_\ell(S_k^{\EN})}{\PPL_\ell(S_k^{\mathrm{mean}})}.
\label{eq:pplratio}
\end{equation}
$R_{\ell,k}>1$ favors language-balanced raw-mean calibration. Perplexity is compared only within the same model and evaluation language; tokenizer-dependent values are not compared across models or languages.

As a metric robustness check, we also compute relative hidden-state change
\begin{equation}
\Rtwo_l^{(\lambda)}
=\mathbb{E}_{x,t}
\left[
\frac{\|\mathbf{h}_{l,t}^{\mathrm{out}}-\mathbf{h}_{l,t}^{\mathrm{in}}\|_2}
{\|\mathbf{h}_{l,t}^{\mathrm{in}}\|_2}
\right],
\end{equation}
but BI determines every primary pruning set.

\section{Experimental Setup}
\label{sec:setup}

\subsection{Models and checkpoint provenance}

Table~\ref{tab:models} covers three decoder-only architecture families in the 1--3B parameter range \citep{qwen2025technical,dubey2024llama,gemma2024report}. We use the organization-hosted Hugging Face checkpoints \code{Qwen/Qwen2.5-1.5B-Instruct}, \code{meta-llama/Llama-3.2-1B}, and \code{google/gemma-2-2b}. Every reported result is tied to an immutable repository revision listed in Appendix~\ref{app:reproducibility}.

\begin{table}[t]
\caption{Models and pruning fractions. Fractions count decoder blocks, not parameters.}
\label{tab:models}
\centering
\small
\begin{tabular}{lrrccc}
\toprule
Model & Parameters & Blocks & $k=2$ & $k=4$ & $k=6$ \\
\midrule
Qwen2.5-1.5B-Instruct & 1.54B & 28 & 7.1\% & 14.3\% & 21.4\% \\
Llama-3.2-1B & 1.24B & 16 & 12.5\% & 25.0\% & 37.5\% \\
Gemma-2-2B & 2.61B & 26 & 7.7\% & 15.4\% & 23.1\% \\
\bottomrule
\end{tabular}
\end{table}

This difference in pruning fraction is important: equal $k$ does not imply equal severity for Llama. We retain $k\in\{2,4,6\}$ because the scientific object is the discrete selector at fixed $k$, but we avoid interpreting cross-model degradation magnitudes as a model-quality ranking.

\subsection{Languages and data}

We use the 2023-11-01 Wikipedia snapshots for English, Hindi, Marathi, and Tamil \citep{wikimedia2023}. Articles are drawn from a deterministically shuffled stream (seed 42), filtered to contain at least 300 whitespace-delimited words, and tokenized separately for each model. At most one 256-token chunk is retained from any article. For each language, 200 articles form the evaluation set and a disjoint 200 articles form the calibration pool. The article identifiers are identical across all three model runs, and assertions check that no evaluation article appears in calibration.

Models defining a beginning-of-sequence token receive it, leaving 255 text tokens; Qwen receives 256 text tokens. All evaluated chunks contain 255 next-token predictions. The main evaluation therefore contains 200 independent article-level NLL observations per language and model. Baseline bits per character and tokenizer fertility are reported in Appendix~\ref{app:additional}.

\subsection{Selection and controls}

For each of five calibration seeds, a monolingual condition samples 150 of its 200 pool chunks without replacement. A balanced condition samples 50 chunks independently from each of \EN, \HI, and \MR. For every condition, we remove $k\in\{2,4,6\}$ blocks and evaluate the resulting model on all four languages. The primary design has seven conditions (four monolingual and three balanced), five seeds, three pruning depths, and four evaluation languages, yielding 420 observations per model.

We additionally evaluate ten deterministic random block sets at every $k$. Full-pool rankings using all 200 calibration chunks are retained as descriptive diagnostics, not mixed into the equal-budget primary contrasts. No fine-tuning, distillation, or post-pruning recovery is performed.

\subsection{Uncertainty and implementation}

For Eq.~\eqref{eq:latecontrast}, we independently bootstrap the 200 English and 200 target-language calibration articles 5,000 times. For pruning contrasts, we form paired per-article NLL differences and use a hierarchical percentile bootstrap that resamples both the five calibration seeds and the 200 evaluation articles. We present these 95\% intervals as the primary uncertainty summary. Exact layer-set equality is reported separately because a zero contrast can arise mechanically when two selectors choose the same blocks.

All experiments ran in bfloat16 on an NVIDIA A100-SXM4-40GB with Transformers 4.56.2 and PyTorch 2.11.0+cu130. Each pruning evaluation starts from the same in-memory block list, bypasses the selected blocks, and restores the complete model afterward. A final baseline re-evaluation checks restoration. \ifpreprint The public code repository (Appendix~\ref{app:reproducibility}) includes\fi{} configurations, article identifiers, block scores, selected sets, per-seed outcomes, contrasts, and random controls.

\section{Results}
\label{sec:results}

\subsection{Global ranking agreement hides pruning-set disagreement}

\begin{figure}[t]
\centering
\includegraphics[width=\linewidth]{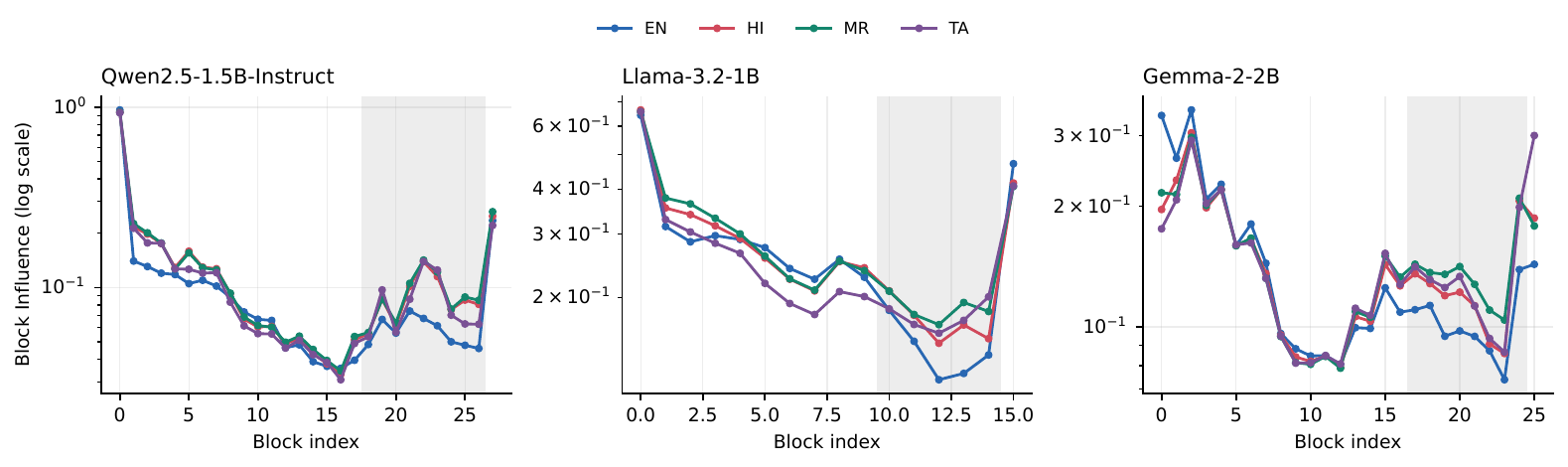}
\caption{Language-conditioned Block Influence. The gray span is the predeclared late region; the final output-adjacent block is excluded from regional contrasts. Curves share a coarse shape but separate near the low-BI late blocks that can enter the pruning set.}
\label{fig:profiles}
\end{figure}

The broad depth profile is shared across languages (Figure~\ref{fig:profiles}). English-to-target Spearman correlations over all BI scores range from 0.898 to 0.982. This apparently strong agreement is not sufficient for pruning: selection depends only on the lowest-scoring tail. Full-pool Jaccard overlap between English and a target-language set falls to 0.33 for Llama and Gemma at $k=2$, 0.60 for Qwen and Gemma at $k=4$, and 0.71 for several $k=6$ comparisons.

The disagreements are concrete. At $k=6$, English calibration for Qwen selects late block 26 while each Indian-language full-pool ranking selects block 13 instead. For Gemma, English selects block 23 at $k=2$ while Hindi and Marathi select block 10. On Llama at $k=2$, English chooses $\{12,13\}$, Hindi chooses $\{12,14\}$, and Marathi and Tamil choose $\{11,12\}$. Thus global rank correlation obscures changes in the exact discrete action.

\subsection{English underestimates late-block influence}

\begin{figure}[t]
\centering
\includegraphics[width=0.92\linewidth]{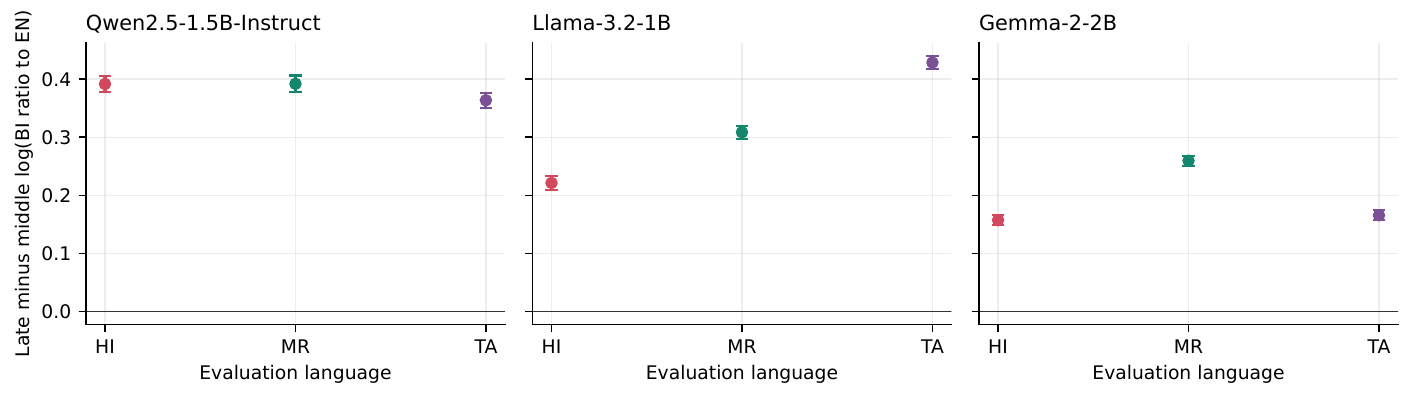}
\caption{Late-minus-middle log BI ratio to English, with 95\% bootstrap intervals. All nine contrasts are above zero.}
\label{fig:late-contrast}
\end{figure}

Table~\ref{tab:late-shift} reports the predeclared depth contrast from Eq.~\eqref{eq:latecontrast}. In the late region, target-language BI is geometrically 20.7--55.5\% higher than English BI across all nine model--language pairs. More importantly, the late-minus-middle contrast is positive in every pair, with every 95\% interval above zero (Figure~\ref{fig:late-contrast}). The smallest contrast is Gemma--Hindi, 0.157 [0.148, 0.166], and the largest is Llama--Tamil, 0.428 [0.417, 0.439].

\begin{table}[t]
\caption{Target-language BI relative to English in the predeclared late region. The late--middle column is Eq.~\eqref{eq:latecontrast}; intervals bootstrap independent calibration articles.}
\label{tab:late-shift}
\centering
\small
\begin{tabular}{llrrr}
\toprule
Model & Language & Late shift (\%) & Late--middle & 95\% CI \\
\midrule
Qwen2.5-1.5B-Instruct & HI & 51.0 & 0.391 & [0.378, 0.405] \\
 & MR & 55.5 & 0.392 & [0.378, 0.406] \\
 & TA & 39.6 & 0.363 & [0.350, 0.376] \\
\addlinespace
Llama-3.2-1B & HI & 20.7 & 0.221 & [0.210, 0.233] \\
 & MR & 31.8 & 0.308 & [0.297, 0.319] \\
 & TA & 25.3 & 0.428 & [0.417, 0.439] \\
\addlinespace
Gemma-2-2B & HI & 21.4 & 0.157 & [0.148, 0.166] \\
 & MR & 36.1 & 0.260 & [0.251, 0.268] \\
 & TA & 24.5 & 0.166 & [0.157, 0.174] \\
\addlinespace
\bottomrule
\end{tabular}

\end{table}

Tamil follows the same pattern despite being non-Devanagari and from a different language family. The result therefore cannot be explained solely as agreement between Hindi and Marathi, although this experiment does not isolate script from every other linguistic or corpus property. This consistent localization is the strongest evidence for our main claim.

\subsection{Balanced raw-mean calibration often helps, but not universally}

\begin{figure}[t]
\centering
\includegraphics[width=\linewidth]{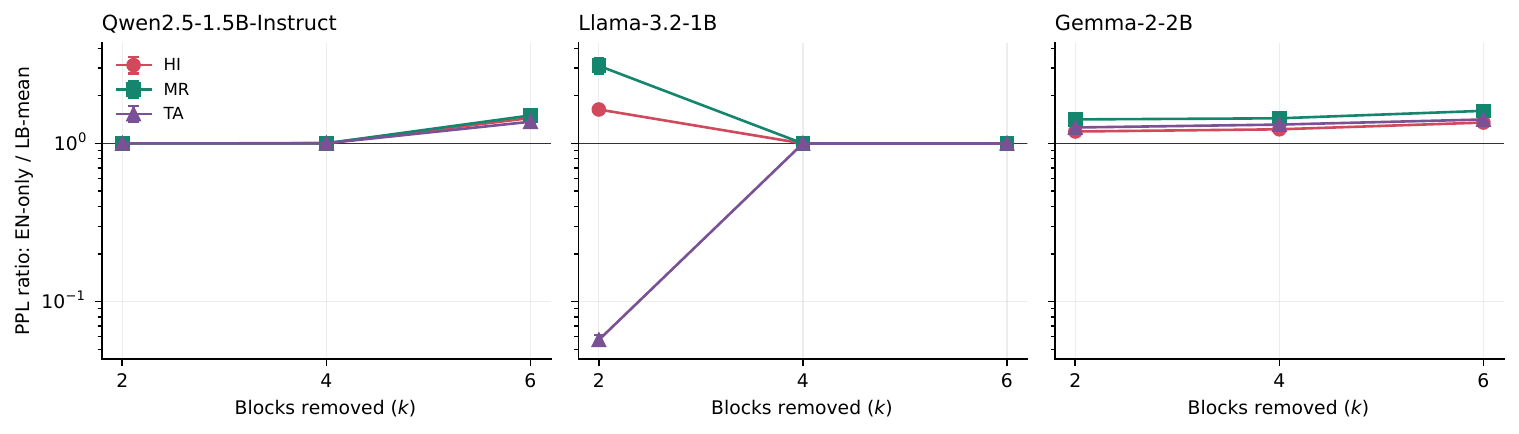}
\caption{English-only versus raw-mean balanced calibration. Values above one favor balanced selection; values below one favor English-only selection. The log scale exposes the Llama--Tamil reversal.}
\label{fig:en-vs-lb}
\end{figure}

Table~\ref{tab:en-vs-lb} and Figure~\ref{fig:en-vs-lb} compare equal-budget English-only selection with raw-mean \EN/\HI/\MR{} selection. Ratios above one favor balancing.

\begin{table}[t]
\caption{Perplexity ratio $R_{\ell,k}=\PPL(S_k^{\EN})/\PPL(S_k^{\mathrm{mean}})$ with hierarchical 95\% intervals. The marker \texttt{=} denotes identical layer sets for all five seeds. Tamil is held out from balanced selector construction.}
\label{tab:en-vs-lb}
\centering
\scriptsize
\resizebox{\linewidth}{!}{\begin{tabular}{llrrr}
\toprule
Model & $k$ & HI & MR & TA \\
\midrule
Qwen2.5-1.5B-Instruct & 2 & 1.00 (=) & 1.00 (=) & 1.00 (=) \\
 & 4 & 1.004 [1.000, 1.012] & 1.003 [1.000, 1.009] & 1.003 [1.000, 1.008] \\
 & 6 & 1.449 [1.434, 1.465] & 1.504 [1.484, 1.522] & 1.372 [1.361, 1.384] \\
\addlinespace
Llama-3.2-1B & 2 & 1.640 [1.553, 1.730] & 3.098 [2.768, 3.487] & 0.057 [0.053, 0.061] \\
 & 4 & 1.00 (=) & 1.00 (=) & 1.00 (=) \\
 & 6 & 1.00 (=) & 1.00 (=) & 1.00 (=) \\
\addlinespace
Gemma-2-2B & 2 & 1.190 [1.178, 1.201] & 1.419 [1.402, 1.437] & 1.262 [1.248, 1.276] \\
 & 4 & 1.229 [1.219, 1.239] & 1.442 [1.428, 1.457] & 1.316 [1.302, 1.332] \\
 & 6 & 1.356 [1.336, 1.375] & 1.608 [1.581, 1.635] & 1.417 [1.394, 1.441] \\
\addlinespace
\bottomrule
\end{tabular}
}
\end{table}

\paragraph{Qwen.}
At $k=2$, both selectors choose the same blocks. At $k=4$, they differ for only one of five calibration seeds and the effect is small, so we treat this setting as inconclusive. At $k=6$, balancing lowers PPL in all three target languages: English-only PPL is $1.45\times$, $1.50\times$, and $1.37\times$ balanced PPL on Hindi, Marathi, and Tamil. Mean target-language degradation falls from 88.9\% to 30.8\%, and worst-language degradation from 107.5\% to 38.0\%.

\paragraph{Gemma.}
Raw-mean balancing changes the selected set at all three depths and improves every target language. English-to-balanced PPL ratios range from $1.19$ to $1.61$. Mean target degradation falls from 50.8\% to 16.8\% at $k=2$, from 78.5\% to 34.2\% at $k=4$, and from 242.3\% to 132.5\% at $k=6$. Held-out Tamil improves at every depth, showing that the effect is not limited to languages included in the average.

\paragraph{Llama is a counterexample to universal robustness.}
At $k=2$, balancing changes $\{12,13\}$ to $\{12,14\}$ and improves Hindi and Marathi: English-only PPL is $1.64\times$ and $3.10\times$ balanced PPL. The same change is disastrous for held-out Tamil, where balanced PPL is approximately $17.5\times$ English-only PPL. Consequently, balanced target-mean degradation is 2694\%, versus 1462\% for English, and its worst-language degradation is 6572\%, versus 3081\%. At $k=4$ and $k=6$, the selectors choose identical blocks and both cause extreme degradation. Language balancing therefore corrects two observed languages without producing a robust cross-language choice on this model.

\subsection{Aggregator ablation}
\label{sec:aggregators}

No rank-based rule dominates. For Qwen, rank mean has the lowest target mean at $k=4$ (18.9\% versus 20.4\% for raw mean), while raw mean and rank mean tie at $k=6$ (30.8\%). For Gemma, all three aggregators tie at $k=2$, but only raw mean avoids the English-selected late blocks at $k=4$ and $k=6$; rank mean and minimax rank reproduce the worse English set. All three aggregators select the same blocks on Llama. The complete mean/worst table is in Appendix~\ref{app:additional}.

Raw mean is consequently the most consistent of the tested aggregators, not a universally optimal method. Minimax rank's appealing requirement that a block rank low in every language does not translate into better observed outcomes here.

\subsection{Stability and random controls}

At the 150-chunk budget, all Llama and Gemma selectors have pairwise Jaccard 1.0 across five seeds. Qwen's minimum is 0.84, and its full-pool agreement minimum is 0.68. English-only selections are themselves perfectly stable in these runs. We therefore use \emph{robust} only for cross-language outcome, not as a synonym for calibration-seed stability.

For Qwen and Gemma, BI-selected sets have lower average target degradation than the ten random sets at every $k$. Llama again exposes a method-level limitation: both English BI and balanced BI are worse than the random-set average at $k=2$ and $k=4$, although BI is better at $k=6$. This result, together with Llama's high pruning fractions in Table~\ref{tab:models}, shows that a corrected calibration distribution cannot rescue an unreliable importance proxy or an excessively aggressive depth reduction.

\section{Discussion}
\label{sec:discussion}

\paragraph{What the evidence establishes.}
The result that survives every model and target language is not that one aggregation rule always wins. It is that English changes the BI profile selectively with depth: the target-to-English ratio is consistently more positive late than in the middle. This is visible for Hindi, Marathi, and Tamil, with narrow article-bootstrap intervals. English calibration can therefore make a block look relatively inactive precisely where another language assigns it more representational change.

\paragraph{What remains an interpretation.}
A plausible account is that middle blocks perform more language-shared computation while later blocks participate in language-specific surface realization, consistent with prior representation analyses \citep{tang2024neurons,wendler2024llamas,zhao2024multilingualism}. Our experiment does not intervene on linguistic features and therefore does not prove this mechanism. BI is also not causal importance: it measures local angular change, while downstream damage depends on composition with every subsequent block.

\paragraph{Detection and correction are distinct.}
Qwen and Gemma connect the late-layer mismatch to consequential English-selected blocks, and raw-mean balancing removes those choices. Llama retains the same directional mismatch yet defeats the proposed correction on held-out Tamil. This distinction is scientifically useful. A calibration diagnostic can be reproducible even when a selection heuristic is too coarse, the model is too shallow, or the requested pruning fraction is too aggressive.

\paragraph{Practical implications.}
For multilingual BI pruning, we recommend reporting per-language score profiles and the exact selected sets; testing at least one language excluded from selector construction; keeping calibration token budgets equal; comparing against random block sets; and treating the number of removed blocks as a fraction of model depth. Raw-mean balancing is a reasonable candidate selector in our experiments, especially because it retains magnitude information that the rank aggregators discard, but it should be validated on the intended deployment languages and model.

\paragraph{Why no recovery stage is included.}
Recovery training can make a pruned model more useful, but it also introduces a second language distribution, an optimizer, and additional data. The present question is about the initial BI decision. Holding all remaining weights fixed isolates that decision and exposes failures that adaptation could conceal. Our conclusions therefore apply to one-shot selection and immediate language-modeling behavior; they neither support nor rule out language-dependent effects after recovery.

\section{Limitations and Threats to Validity}
\label{sec:limitations}

\paragraph{Model and pruning scale.}
We study three models between 1.24B and 2.61B parameters. Larger or more multilingual models may distribute language-specific computation differently. Equal $k$ removes a larger depth fraction from 16-block Llama than from Qwen or Gemma; this likely contributes to Llama's collapse. The paper reports those fractions and does not use absolute cross-model degradation as a comparison of model quality.

\paragraph{Language coverage.}
Hindi and Marathi are related and share Devanagari. Tamil adds a Dravidian, non-Devanagari control, but all three target languages are Indian languages and use non-Latin scripts. Without a Latin-script non-English control such as Spanish or German, the experiment cannot fully separate an Indian-language effect from a broader English-versus-non-English effect. We consequently phrase the finding for the languages studied rather than for all multilingual text.

\paragraph{Data and outcomes.}
Wikipedia is a single encyclopedic domain, and editions differ in size, style, and quality. The primary outcome is held-out next-token perplexity. It directly measures the language-modeling consequence of block removal but does not guarantee corresponding changes in generation quality, factuality, or downstream-task accuracy. We do not study recovered models, instruction tasks, latency, or energy.

\paragraph{Importance proxy.}
BI captures local angular change and can fail as a causal measure of block necessity. Llama's performance relative to random pruning demonstrates this limitation directly. Relative-$L_2$ profiles are supplied as a diagnostic, but all primary selections use BI.

\paragraph{Sampling and inference.}
The study uses one deterministic Wikipedia stream, 200 calibration and 200 evaluation articles per language, and five calibration subsets. Hierarchical intervals represent these calibration-subset and article sources of variation, not alternative corpora or model checkpoints. Discrete layer selection also produces repeated outcomes and zero seed variance; this should not be read as zero evaluation uncertainty.

\paragraph{Checkpoint provenance.}
All reported experiments use organization-hosted repositories, but the conclusions remain specific to the exact checkpoint revisions listed in Appendix~\ref{app:reproducibility}. Other releases, quantizations, or instruction-tuned variants may yield different block rankings and pruning outcomes.

\paragraph{Exploratory scope.}
The early/middle/late partition and primary budget were fixed in code, but this was not a formally registered study. The evidence should be replicated on additional domains, Latin-script non-English languages, and larger models before stronger population-level claims are made.

\section{Broader Impact}

Multilingual compression can reduce serving cost and make language technology more accessible under limited hardware budgets. It can also silently shift quality loss toward languages that are absent from calibration. The main positive implication of this work is procedural: evaluate block-selection decisions on the languages a system is expected to serve, including languages not used to construct the selector.

The results should not be used to certify a pruned model as safe for a language. We evaluate language modeling on Wikipedia, not harmful generation, dialectal coverage, or application-specific reliability. A deployment should retain task-specific and human evaluation, document its calibration mixture, and avoid using average multilingual performance to conceal a severe failure in one language.

\section{Conclusion}

English-only calibration systematically assigns lower relative BI to late blocks than Hindi, Marathi, and Tamil across Qwen2.5-1.5B, Llama-3.2-1B, and Gemma-2-2B. The effect is localized, statistically consistent over held-out calibration articles, and large enough to change which blocks are removed despite high global rank correlations.

Equal-budget raw-mean calibration over English, Hindi, and Marathi avoids damaging English-selected blocks for Qwen at the largest tested budget and for Gemma at every tested budget, including gains on held-out Tamil. Llama prevents a broader claim: balancing helps Hindi and Marathi at two removed blocks but sharply worsens Tamil, and BI pruning itself is brittle. The appropriate conclusion is therefore diagnostic and conditional. Calibration language is part of the BI pruning algorithm, and multilingual balance is a candidate mitigation that must be tested per model, pruning fraction, and deployment language.

\section*{Code and Data Availability}

Code, experiment configurations, article identifiers, block scores, pruning sets, and per-seed results are available at \coderepo.

\bibliography{references}
\bibliographystyle{tmlr}

\appendix
\section{Reproducibility and Provenance}
\label{app:reproducibility}

\subsection{Exact checkpoints}

\begin{table}[h]
\caption{Model identifiers and immutable repository revisions used for the reported runs.}
\label{tab:checkpoints}
\centering
\scriptsize
\begin{tabular}{lll}
\toprule
Model & Repository identifier & Revision \\
\midrule
Qwen & \code{Qwen/Qwen2.5-1.5B-Instruct} & \code{989aa7980e4c...} \\
Llama & \code{meta-llama/Llama-3.2-1B} & \code{4e20de362430...} \\
Gemma & \code{google/gemma-2-2b} & \code{c5ebcd40d208...} \\
\bottomrule
\end{tabular}
\end{table}

The complete 40-character revisions are stored in each \ifpreprint released\fi{} \code{experiment\_config.json}. All three identifiers point to repositories maintained by the respective model organizations. Each run used a model-specific results directory and recorded the model identifier, immutable revision, software environment, numerical precision, evaluation fingerprint, and cache statistics before its outputs were incorporated into the paper.

\subsection{Data construction}

The sampling procedure is deterministic given seed 42. The cached source stream is shuffled with buffer size 5,000. For each language, the first 200 valid articles provide evaluation chunks and the next 200 provide calibration chunks. The code stores all article identifiers and asserts uniqueness and cross-split disjointness. Cross-model asset generation additionally asserts equality of the article-identifier tables before combining any results.

Every calibration and evaluation row corresponds to one article. The independent article, rather than a token or overlapping window, is therefore the resampling unit. The paper artifact includes \code{calibration\_article\_ids.csv} and \code{evaluation\_article\_ids.csv} for every model.

\subsection{Artifact map}

\ifpreprint
The public repository at \coderepo{} contains:
\fi
\begin{itemize}
    \item the complete v4 experiment notebook and its deterministic notebook builder;
    \item exact model configurations and revisions;
    \item per-layer BI and relative-$L_2$ arrays summarized by language;
    \item full-pool and per-seed pruning sets;
    \item all 420 primary rows and 99 contrast rows per model;
    \item random-pruning controls, stability summaries, and article identifiers; and
    \item a SHA-256 manifest for every compact result file.
\end{itemize}

The large model weights and per-configuration logits are not redistributed. \ifpreprint Per-chunk NLL caches are also not redistributed.\fi{} The included tables are sufficient to verify every number in the manuscript.

\subsection{Software and execution}

The recorded environment is Transformers 4.56.2, PyTorch 2.11.0+cu130, bfloat16, and one NVIDIA A100-SXM4-40GB. The notebook records 200 influence chunks and 200 evaluation chunks per language, sequence length 256, five calibration seeds, primary budget 150, and ten random seeds. Optional budget-sweep and Belebele cells were disabled; neither contributes to any claim in this paper.

\section{Additional Results}
\label{app:additional}

\subsection{Baseline quality and tokenization}

\begin{table}[h]
\caption{Intact-model perplexity (PPL), bits per character (BPC), and mean tokenizer tokens per whitespace-delimited word. Values are descriptive; PPL is not compared across tokenizers.}
\label{tab:baselines}
\centering
\scriptsize
\begin{tabular}{llrrr}
\toprule
Model & Language & PPL & BPC & Tokens/word \\
\midrule
\multirow{4}{*}{Qwen2.5-1.5B} & EN & 10.207 & 0.790 & 1.533 \\
 & HI & 3.827 & 1.776 & 4.996 \\
 & MR & 5.256 & 2.284 & 6.730 \\
 & TA & 3.620 & 1.920 & 9.553 \\
\midrule
\multirow{4}{*}{Llama-3.2-1B} & EN & 10.158 & 0.755 & 1.459 \\
 & HI & 6.093 & 1.357 & 2.854 \\
 & MR & 6.939 & 1.597 & 4.009 \\
 & TA & 2.321 & 1.507 & 11.635 \\
\midrule
\multirow{4}{*}{Gemma-2-2B} & EN & 8.916 & 0.727 & 1.500 \\
 & HI & 8.879 & 1.262 & 2.210 \\
 & MR & 9.154 & 1.508 & 3.389 \\
 & TA & 6.057 & 1.233 & 4.312 \\
\bottomrule
\end{tabular}
\end{table}

Tamil has the highest tokenizer fertility in each model, yet balanced calibration transfers positively to Tamil for Qwen and Gemma and negatively for Llama. Fertility alone therefore does not explain the direction of the mitigation result.

\subsection{Complete aggregator summary}

Table~\ref{tab:all-strategies} reports mean and worst relative PPL degradation over Hindi, Marathi, and Tamil. Means and standard deviations are over calibration seeds. A standard deviation of zero indicates that all five subsets selected the same block set; it does not remove evaluation-article uncertainty.

\begin{table}[p]
\caption{Target-language relative PPL degradation (\%) for English-only and balanced BI selectors.}
\label{tab:all-strategies}
\centering
\tiny
\resizebox{\linewidth}{!}{\begin{tabular}{lllrrrrrr}
\toprule
Model & Selector & Statistic & $k=2$ & SD & $k=4$ & SD & $k=6$ & SD \\
\midrule
Qwen2.5-1.5B-Instruct & EN-only & Mean & 6.68 & 0.00 & 20.72 & 0.00 & 88.89 & 0.00 \\
 &  & Worst & 8.45 & 0.00 & 25.61 & 0.00 & 107.5 & 0.00 \\
\addlinespace[1pt]
 & Raw mean & Mean & 6.68 & 0.00 & 20.36 & 0.81 & 30.84 & 0.00 \\
 &  & Worst & 8.45 & 0.00 & 25.27 & 0.76 & 38.03 & 0.00 \\
\addlinespace[1pt]
 & Rank mean & Mean & 6.68 & 0.00 & 18.90 & 0.00 & 30.84 & 0.00 \\
 &  & Worst & 8.45 & 0.00 & 23.91 & 0.00 & 38.03 & 0.00 \\
\addlinespace[1pt]
 & Minimax rank & Mean & 6.68 & 0.00 & 19.26 & 0.81 & 34.17 & 7.44 \\
 &  & Worst & 8.45 & 0.00 & 24.25 & 0.76 & 41.58 & 7.93 \\
\addlinespace[1pt]
\midrule
Llama-3.2-1B & EN-only & Mean & 1,462 & 0.00 & 9.06e+05 & 0.00 & 8.76e+05 & 0.00 \\
 &  & Worst & 3,081 & 0.00 & 2.45e+06 & 0.00 & 1.93e+06 & 0.00 \\
\addlinespace[1pt]
 & Raw mean & Mean & 2,694 & 0.00 & 9.06e+05 & 0.00 & 8.76e+05 & 0.00 \\
 &  & Worst & 6,572 & 0.00 & 2.45e+06 & 0.00 & 1.93e+06 & 0.00 \\
\addlinespace[1pt]
 & Rank mean & Mean & 2,694 & 0.00 & 9.06e+05 & 0.00 & 8.76e+05 & 0.00 \\
 &  & Worst & 6,572 & 0.00 & 2.45e+06 & 0.00 & 1.93e+06 & 0.00 \\
\addlinespace[1pt]
 & Minimax rank & Mean & 2,694 & 0.00 & 9.06e+05 & 0.00 & 8.76e+05 & 0.00 \\
 &  & Worst & 6,572 & 0.00 & 2.45e+06 & 0.00 & 1.93e+06 & 0.00 \\
\addlinespace[1pt]
\midrule
Gemma-2-2B & EN-only & Mean & 50.82 & 0.00 & 78.52 & 0.00 & 242.3 & 0.00 \\
 &  & Worst & 68.36 & 0.00 & 97.66 & 0.00 & 327.6 & 0.00 \\
\addlinespace[1pt]
 & Raw mean & Mean & 16.81 & 0.00 & 34.19 & 0.00 & 132.5 & 0.00 \\
 &  & Worst & 18.65 & 0.00 & 37.04 & 0.00 & 165.9 & 0.00 \\
\addlinespace[1pt]
 & Rank mean & Mean & 16.81 & 0.00 & 78.52 & 0.00 & 242.3 & 0.00 \\
 &  & Worst & 18.65 & 0.00 & 97.66 & 0.00 & 327.6 & 0.00 \\
\addlinespace[1pt]
 & Minimax rank & Mean & 16.81 & 0.00 & 78.52 & 0.00 & 242.3 & 0.00 \\
 &  & Worst & 18.65 & 0.00 & 97.66 & 0.00 & 327.6 & 0.00 \\
\addlinespace[1pt]
\addlinespace
\bottomrule
\end{tabular}
}
\end{table}
\clearpage

\subsection{Full-pool selection examples}

\begin{table}[h]
\caption{Full-pool BI pruning sets for the two primary selectors. Indices are zero-based.}
\label{tab:key-sets}
\centering
\scriptsize
\begin{tabular}{llccc}
\toprule
Model & Selector & $k=2$ & $k=4$ & $k=6$ \\
\midrule
\multirow{2}{*}{Qwen} & EN & 15,16 & 14--17 & 12,14--17,26 \\
 & Raw mean & 15,16 & 14--17 & 12--17 \\
\midrule
\multirow{2}{*}{Llama} & EN & 12,13 & 11--14 & 7,10--14 \\
 & Raw mean & 12,14 & 11--14 & 7,10--14 \\
\midrule
\multirow{2}{*}{Gemma} & EN & 12,23 & 10--12,23 & 9--12,22,23 \\
 & Raw mean & 10,12 & 9--12 & 8--12,23 \\
\bottomrule
\end{tabular}
\end{table}

The complete monolingual and aggregated selections, including all five primary seeds, are supplied as JSON. Full-pool sets are descriptive because the primary equal-budget analysis uses 150-chunk subsets.

\subsection{\texorpdfstring{Rank agreement and relative-$L_2$ diagnostic}{Rank agreement and relative-L2 diagnostic}}

For BI, English-to-target Spearman correlation ranges are 0.898--0.910 (Qwen), 0.947--0.982 (Llama), and 0.906--0.949 (Gemma). Corresponding Jaccard overlaps can be as low as 0.33. Relative-$L_2$ produces the same broad depth shape but lower agreement in several pairs; its complete correlations and per-layer values are supplied. Because the study's hypothesis and all selection outcomes are defined for BI, we do not reinterpret relative-$L_2$ as a second primary test.

\subsection{Random-pruning target means}

\begin{table}[h]
\caption{Mean relative PPL degradation (\%) over Hindi, Marathi, and Tamil. Random is averaged over ten block sets; EN and raw mean are averaged over five calibration seeds.}
\label{tab:random}
\centering
\scriptsize
\begin{tabular}{lrrrrrrrrr}
\toprule
& \multicolumn{3}{c}{Qwen} & \multicolumn{3}{c}{Llama} & \multicolumn{3}{c}{Gemma} \\
\cmidrule(lr){2-4}\cmidrule(lr){5-7}\cmidrule(l){8-10}
$k$ & EN & LB & Random & EN & LB & Random & EN & LB & Random \\
\midrule
2 & 6.7 & 6.7 & 72.8 & 1461.7 & 2694.4 & 429.7 & 50.8 & 16.8 & 198.6 \\
4 & 20.7 & 20.4 & 1122.1 & $9.06{\times}10^5$ & $9.06{\times}10^5$ & $3.13{\times}10^5$ & 78.5 & 34.2 & $7.48{\times}10^4$ \\
6 & 88.9 & 30.8 & $1.81{\times}10^7$ & $8.76{\times}10^5$ & $8.76{\times}10^5$ & $1.01{\times}10^7$ & 242.3 & 132.5 & $2.11{\times}10^8$ \\
\bottomrule
\end{tabular}
\end{table}

\section{Statistical Details}
\label{app:statistics}

\subsection{Late-layer bootstrap}

For each target language and model, we draw 200 English calibration chunks and 200 target-language chunks independently with replacement. BI is recomputed by averaging the stored chunk-level influence values, then transformed into layer-wise log ratios using Eq.~\eqref{eq:logratio}. The middle and late means and their difference are calculated for each of 5,000 replicates. Table~\ref{tab:late-shift} uses percentile intervals from this distribution.

\subsection{Hierarchical pruning contrast}

For conditions $a$ and $b$, seed $s$, and evaluation article $i$, let
\[
d_{s,i}=\operatorname{NLL}_{a,s,i}-\operatorname{NLL}_{b,s,i}.
\]
Each bootstrap replicate samples five seed indices with replacement and one shared set of 200 article indices with replacement. It averages the resulting $5\times200$ differences. The reported PPL ratio is $\exp(\overline{d})$, and interval endpoints exponentiate the 2.5th and 97.5th percentiles of the log-scale replicates.

This procedure preserves pairing between two pruning conditions on the same evaluation article while representing variation from both calibration subsets and evaluation articles. When conditions select identical sets, all differences are identically zero. When only one seed differs, as for Qwen raw mean at $k=4$, the lower endpoint can touch one; we describe that result as inconclusive rather than using rounded endpoints to claim a reliable gain.

\subsection{Multiplicity and interpretation}

The artifact includes exploratory sign-flip tests with Holm adjustments because they were generated by the experiment notebook. The paper does not use those $p$-values as its primary evidence: with only five calibration subsets and a discrete selector, averaging seeds before a sign-flip test can underrepresent selector variation. We instead report effect sizes, hierarchical intervals, exact set-equality rates, and consistency of the predeclared late-depth pattern.

\section{Implementation Checks}
\label{app:implementation}

\paragraph{Block outputs.}
The final entry returned by a model's global hidden-state API can include the final RMS normalization. Comparing that tensor with the input to the last decoder block inflates the apparent influence of the last block. We use forward hooks on each decoder block and compare its direct input and output.

\paragraph{Beginning-of-sequence handling.}
Llama and Gemma define a beginning-of-sequence token. Omitting it caused implausible Gemma baselines in pilot experiments. We prepend it for models that define it and exclude its position from BI because attention sinks can have outlier activations \citep{xiao2024streaming}.

\paragraph{No cumulative pruning.}
Every evaluated set is applied to the original ordered list of decoder blocks. A context manager restores that list after evaluation, and the notebook re-evaluates the intact baseline at the end. Thus the $k=4$ model is not obtained by further pruning a previously modified $k=2$ model.

\paragraph{Cache identity.}
Per-configuration NLL cache keys include model identifier, repository revision, sequence length, and evaluation-article identifiers. Influence files are kept in model-specific result directories, preventing cached outputs from being reused across checkpoints.

\paragraph{Determinism.}
Block selection uses deterministic tie-breaking. Random controls use fixed seeds derived from $k$. The code records package versions, model revision, all selected sets, and an evaluation fingerprint. Data loading is the only network-dependent stage once model files are available; sampled article records are cached.

\end{document}